\documentclass[conference]{IEEEtran}
\IEEEoverridecommandlockouts
\pdfoutput=1
\usepackage{cite}
\usepackage{amsmath,amssymb,amsfonts}
\usepackage{algorithmic}
\usepackage{graphicx}
\usepackage{textcomp}
\usepackage{xcolor}
\begin{document}

\title{Setup of a Recurrent Neural Network as a Body Model for Solving Inverse and Forward Kinematics as well as Dynamics for a Redundant Manipulator
\thanks{This research was supported by the Cluster of Excellence Cognitive Interaction Technology `CITEC' (EXC 277) at Bielefeld University, which is funded by the German Research Foundation (DFG). }
}

\author{\IEEEauthorblockN{Malte Schilling}
\IEEEauthorblockA{\textit{Center of Excellence `Cognitive Interaction Technology'} \\
Bielefeld University, Germany,\\
mschilli@techfak.uni-bielefeld.de}
}

\maketitle

\begin{abstract}
An internal model of the own body can be assumed a fundamental and evolutionary-early representation as it is present throughout the animal kingdom. Such functional models are, on the one hand, required in motor control, for example solving the inverse kinematic or dynamic task in goal-directed movements or a forward task in ballistic movements. On the other hand, such models are recruited in cognitive tasks as are planning ahead or observation of actions of a conspecific. Here, we present a functional internal body model that is based on the Mean of Multiple Computations principle. For the first time such a model is completely realized in a recurrent neural network as necessary normalization steps are integrated into the neural model itself. Secondly, a dynamic extension is applied to the model. It is shown how the neural network solves a series of inverse tasks. Furthermore, emerging representation in transformational layers are analyzed that show a form of prototypical population-coding as found in place or direction cells.
\end{abstract}


\section{Introduction}
Already simple animals like insects are able to find their way around and to efficiently handle quite complex behaviors like climbing through trees or on small twigs. Usually, such systems are supposed to rely only on very simplistic controller architectures and the control is assumed to not involve detailed knowledge processing or internal modeling of the surroundings. Instead, such systems are situated in the environment and are thought to mainly use embodied principles \cite{brooks}. Animals have inspired robotics for a long time and especially the embodied approach has been highly influenced by biology and tried to incorporate insights from behavioral analysis. As a consequence, bottom-up approaches start with low level behaviors and try to circumvent internal models and representation as far as possible as those are always subject to error. While those approaches have been---and are still---successful, it is also reasonable that higher level function involves forms of internal representation. The bottom-up approach does not exclude such models, but offers a helpful perspective \cite{steels03pt}. Internal models are seen in this view not as a means to themselves or for higher level function, but at first they are connected to (and grounded in) a specific behavior. An internal model serves initially action and has co-evolved with this. Only later-on it might be utilized in more flexible ways.

Usually, one distinguishes different types or function of such internal models in the context of action. In particular, there are two types of kinematic models. First, inverse models deal with transformations from a Cartesian three dimensional space---as, for example, visually perceived---into an actuator or action space like joint values or muscle activations \cite{wolpert98modules}. A typical example can be found in grasping and reaching movements in humans and animals \cite{shadmehr05}. For complex kinematic structures such problems become quite hard or intractable. In particular, for redundant manipulators there are more degrees of freedom than required and as a consequence there are often multiple possible solutions. Secondly, forward models deal with the inverse transformation. A forward model predicts a position in space when given current joint values or muscles activations. Such models are, for example, used in fast and ballistic reaching movements for which sensory feedback is too slow to modulate motor control and instead it is assumed that a predicted outcome of the movement is utilized \cite{wolpert01}. Such simple internal models can be found throughout the animal kingdom, even in insects \cite{menzel07invertebrate,durr2018transfer}. Importantly, such internal models are also used in different contexts and allow for cognitive capabilities, like planning ahead \cite{schilling2017reacog}. The ability to plan ahead is crucial for cognition \cite{mcfarland93cognitive} and is thought to be realized as a form of internal simulation that is applied using such flexible internal models. In particular, forward models are of predictive nature which allows to recruit these predictive capabilities for planning ahead in an internal simulation \cite{jeannerod99act}.

In motor control it is usually distinguished between inverse and forward functions. This leads to very specific internal models serving one specific behavior and one specific function. A good example is the influential MOSAIC approach \cite{wolpert98modules}, in which each behavior has a pair of inverse and forward models. Such an approach has some drawbacks, as it is not very efficient and as there are duplicate representations. Adaptation of such models seems quite problematic---when the body changes it has to alter all individual models. Therefore, there is the longstanding notion of a single internal body representation \cite{acost05body, hoffmann2010bodymodel} that might subserve these functions in the context of different behaviors and tasks.

There is a large body of literature on how such a body schema might be neuronally encoded in humans and animals. In general, it is assumed that configurations of body parts are encoded in a distributed and somatotopic fashion in distinct areas in the brain and that there are redundant representations \cite{Andersen532, Georgopoulos2928}. Recent findings provide further information on the organization and forms of encoding of such internal body models. Mimica et al. \cite{mimica_efficient_2018} identified topographically organized populations of neurons that encode the posture in the posterior parietal cortex and frontal motor cortex of freely moving rats. Neurons in these areas are tuned toward specific postures of body parts and appear to encode configuration of that body part in a distributed population. It is assumed that these areas and these population serve coordinate transformations as well as integration of sensory signals provided with respect to different frames of reference. This suggests that these neurons realize a form of an internal body model as detailed above and that body part configurations are encoded in these areas in a redundant as well as distributed and population-based fashion. 

In this article, we introduce an internal body model that is fully realized as a simple recurrent neural network. This model is based on the Mean of Multiple Computations (MMC) principle which will be explained in the next section and which allows to address both kinematic problems as mentioned above. While the original MMC model involved external application of non-linear constraints, the novel contribution of this article is the integration of these normalization constraints directly into the neural network model. These normalization sub-networks will be introduced afterwards. As a further step, the model will be extended towards a dynamic model. Last, we present results, first, for the performance of the learned normalization sub-networks and an analysis of the emerging population codes that show prototype-like characteristics as can be found in the biological experiments. This will be followed by results for the overall body model and results for the dynamic model.

\begin{figure}[tbp]
\centerline{\includegraphics[width=0.85\columnwidth]{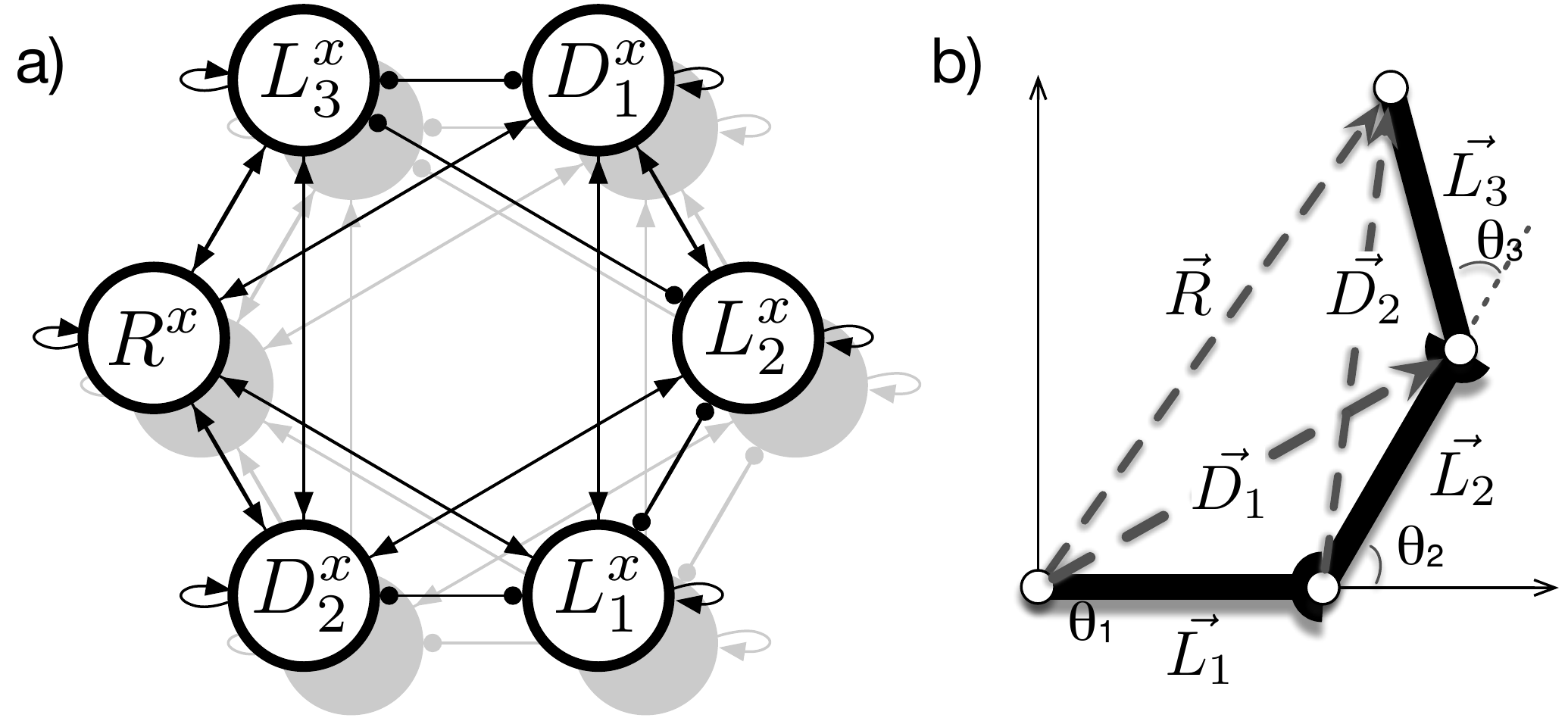}}
\caption{An MMC network as a recurrent neural network representing a three segmented arm. The structure of the manipulator is shown in b). In a) the network connections are visualized which consist of influences between the different variables. Connections are setup following the MMC principle.}
\label{fig_arm}
\end{figure}

\section{The Mean of Multiple Computations Model}
One example for an internal body model is realized by a Mean of Multiple Computations network \cite{schilling2011dq},. 
The general structure of the neural network is not learned, but setup following the MMC principle. The model encodes geometric and kinematic constraints in a recurrent neural network. This shapes an attractor space that represents valid configurations of the encoded manipulator structure. Importantly, diverse tasks can be given to the network which initially act as a disturbance on a previous state of the network, but which are resolved through spreading the disturbance through the network and settling in a new valid state. For example, an inverse kinematic problem can be solved by injecting a new position for the end effector into the network which requires the individual segments of the controlled structure to adapt towards new positions.

The Mean of Multiple Computations principle consists of two key ideas: first, the overall complexity of the controlled structure is broken down. As an example, we will use a three-segmented robotic arm. When the whole kinematic chain is expressed as one single equation, the problem becomes quite difficult for traditional control approaches. In the inverse kinematic task there are multiple possible configurations in order to reach a certain target position. Instead, the MMC principle breaks this down into multiple computations that consist of local relationships between variables. While the individual equations become trivial to solve (they only consist of three variables), we end up with multiple of such computations. As the second key idea, the MMC principle exploits this redundancy. As each variable appears in multiple computations, it depends on multiple of these equations. The MMC network works in an iterative fashion: an update for a variable is calculated using all the equations that affect this variable. The different multiple computations are integrated towards a new value--- this is realized as a simple weighted mean calculation.

After explanation of the general idea, in the next section the MMC principle will be applied to a simple vector representation as has been done in the classical MMC approach. This will illustrate the general approach, but we will also use this to highlight drawbacks of this approach. Afterwards, these drawbacks will be addressed. First, the non-linearities will be reformulated and handled by a simple hidden layer. Second, the network dynamics will be changed through explicit introduction of velocity equations.

\subsection{Vector Approach}
In our example case of a three-segmented manipulator (see Fig. \ref{fig_arm}) all variables are described by vectors as such a representation is straightforward to use. As variables, there is the position of the end-effector ($\vec{R}$) and each segment is represented by a vector ($\vec{L_i}$). Furthermore, two diagonal vectors are introduced ($\vec{D_i}$). The MMC approach breaks down the overall complexity of the manipulator into simple, local relationships and integrates those in a second step. As an example, we are looking at the first segment. The first segment can be described by two different relations:
\begin{align}
\vec{L}_1 = \vec{D}_1 - \vec{L}_2, \vec{L}_1 = \vec{R} - \vec{D}_2 
\label{equ_all_l1}
\end{align}

Each of these equations corresponds to a closed chain of three vectors (meaning the three contributing vectors add up to zero). In the same way, equations can be easily generated and solved for all the other variables. These equations form the basis for the MMC network---and we can find these influences encoded in the neural network, e.g. see in Fig. \ref{fig_arm} a) the positive influence from $\vec{D}_1$ and the inhibiting/negative influence from $\vec{L}_2$ which is equal to the first equation.

In the second step, the different equations are integrated---using a simple weighted mean calculation. Again, for our first segment there are two equations that have to be integrated. As one further part of the integration, the current value of the variable itself is used weighted with a so-called damping weight $d$ (chosen in our simulations $d=10$). This has shown to stabilize the network \cite{steinkuehler98mmc}. Overall, we get for the first segment as an equation:
\begin{align}
\vec{L}_{1}(t+1) = & \frac{1}{d} (\vec{D}_{1}(t) - \vec{L}_{2}(t)) +  \frac{1}{d} (\vec{R}(t) - \vec{D}_{2}(t))  \nonumber \\
&+ \frac{d-2}{d} \vec{L}_{1}(t)
\label{equ_mmc_l1}
\end{align}

\begin{figure}[tbp]
\centerline{\includegraphics[width=0.6\columnwidth]{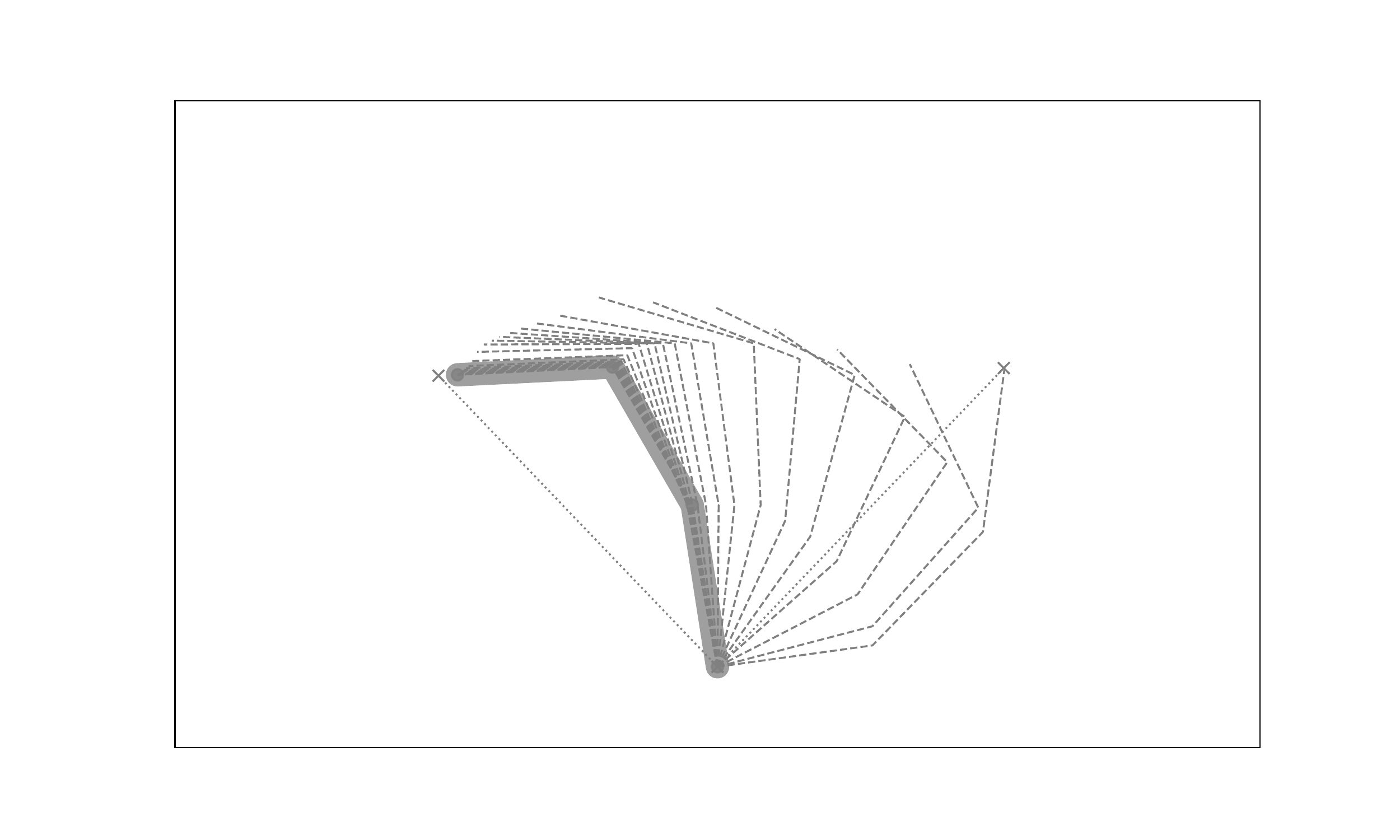}}
\caption{Example solution to an inverse kinematic problem. Shown is the arm at different points in time. The base of the arm is down in the middle and a new target position at the upper left is provided. This changed vector does not fit to the current activations of the other variables, but influences these other variables. As the overall attractor space is constrained through the encoded kinematics, this forces the net to settle into a new state. The arm at the end of the movement is shown as the thick, solid line close to the target. The dashed lines show in between states during the movement---the initial configuration is shown by the dashed lines most to the right.}
\label{fig_example_movement}
\end{figure}

Together, the equations for the different variables can be interpreted as a recurrent neural network as shown in Fig. \ref{fig_arm}. The connections describe how the different variables are used in the computation of the other variables and the network is updated in an iterative fashion. Importantly, there are separate networks for each dimension of the vectors (we end up with one network for the $x$-dimension and one for the $y$-dimension). Such networks can be applied for inverse (Fig.  \ref{fig_example_movement}) and forward kinematic problems \cite{schilling2012s}.

While an MMC network provides a simple solution to kinematic problems that can easily be extended towards much more complex structures \cite{schilling2007hierarchical}, there are two major drawbacks. First, all variables are handled the same way. As a result, when the network updates its variables, for example towards a new target position, all the connected variables get assigned new values. But in the case of the segment vectors, these vectors can not be simple changed in any way---the segments are rigid structures that can only change their orientation, but not their length. The network is not aware of these constraints which have to be enforced. Up until now, this lead to a second processing step in which non-linear constraints were applied to the segment vectors. The novel contribution of this article is introducing a simple extension that incorporates these non-linear constraints into the neural network and makes this additional step unnecessary.

A second drawback concerns the dynamics of the network: initially, the network produces very fast movements that then subsequently slow down. As one solution, dynamic MMC networks had been proposed that alleviated this problem, but could lead to overshooting \cite{schilling09dynamics}. Here, as a second contribution, we adapt this principle leading to more realistic movement profiles without any overshoot.

\subsection{Learning Non-Linearities}\label{sect_norm_net}
At the core of an MMC network is the recurrent neural network introduced above. For this set of linear equations convergence towards optimal solutions can be proven \cite{steinkuehler98mmc}. But application of such a vector-based network requires one further processing step. In order to guarantee the lengths of the actuator segments, these constraints have to be enforced externally and the values have to be adapted. The variables encoding a segment length are normalized outside of the network. This is not biologically plausible and it introduces non-linearities as well as another external processing step that makes the analysis of the network much more difficult. Therefore, we will introduce how this normalization step can be integrated into the recurrent neural network structure.

\begin{figure}[tbp]
\centerline{\includegraphics[scale=0.45]{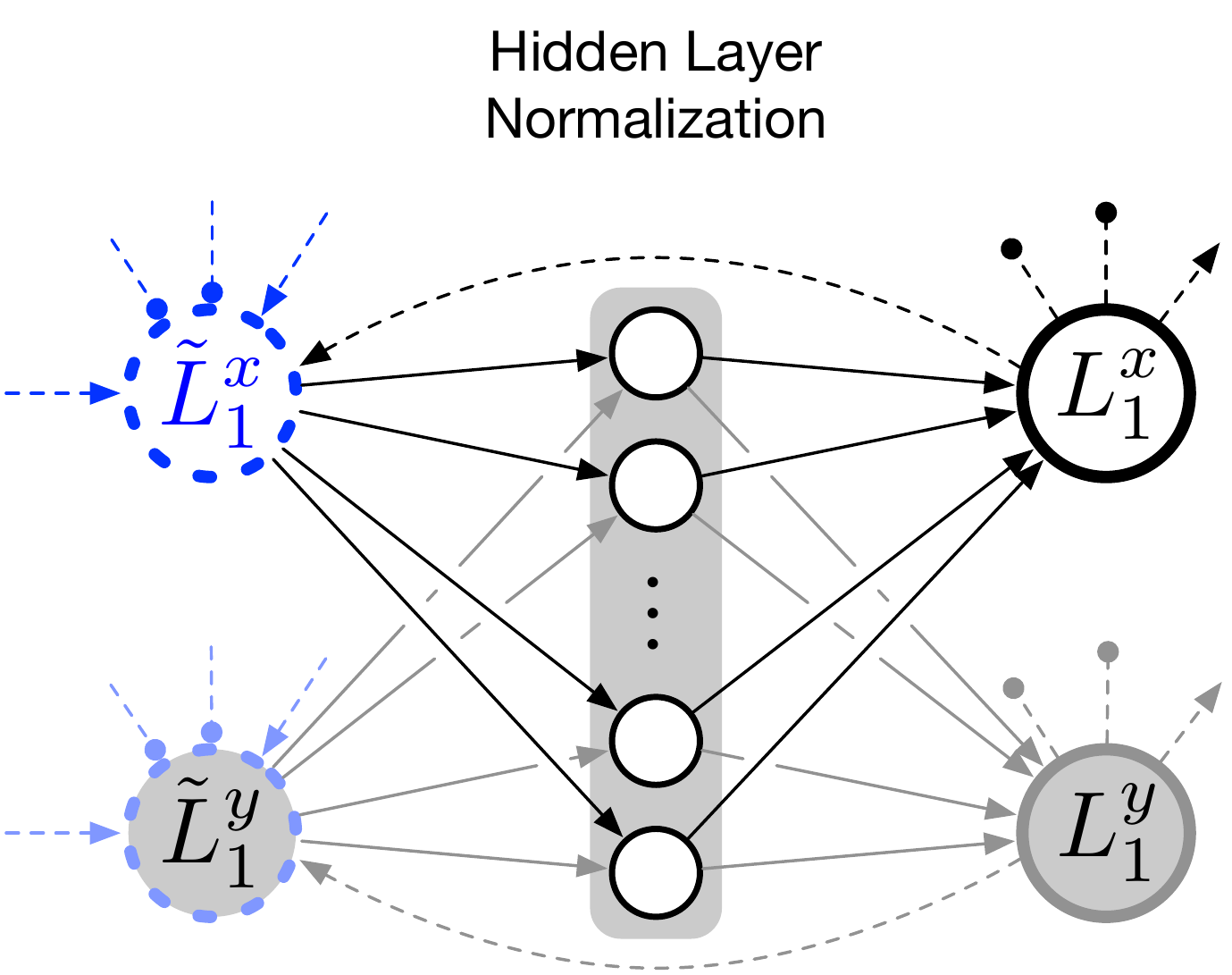}}
\caption{Part of the new MMC network structure: shown is the integrated normalization network for the first segment variable. The process flow inside the network is mainly from left to right: first, the auxiliary variables are calculated following the MMC principle (shown in blue); second, the normalization is performed using the hidden layer in the middle; third, the segment variables are updated. This closes the loop and these are used for outgoing connections inside the recurrent neural network in order to compute the equations.}
\label{fig_hidden_norm}
\end{figure}

In the classical MMC approach, normalization was realized straightforward: x and y values of a variable were taken from the network and the length of this vector was calculated using the Euclidean norm. With a given fixed length of a segment, this allowed to scale the vector accordingly and, in a last step, use these new scaled values to overwrite the variables in the MMC network before initiating the next iteration step. The calculation of the normalization in itself is not difficult and will be now realized through a simple feedforward neural network.

For the proposed model, a simple MLP is introduced that realizes normalization. In a first step, such a model is trained on the normalization task. As an input, random vectors are given to this sub-network and, as an output, the network is expected to provide vectors with the same orientation, but unit length. The network is trained in a supervised manner and already a simple structure of only a couple of hidden units is able to successfully solve the normalization task. 

Secondly, the feedforward normalization sub-net is introduced for all three segment variables into our MMC network and replaces the external constraint application (Fig. \ref{fig_hidden_norm}). Importantly, this requires to introduce auxiliary variables for the segments: on the one hand, there are still the values ($\vec{L_i}$) used for the computations of all the other variables and following the connections inside the MMC network. On the other hand, the new values stemming from the integration of multiple computations have to be buffered in a variable before the normalization step ($\vec{\tilde {L_i}}$). These two variables are connected through the feedforward normalization sub-network with fixed weights. On the input side of this network, the buffered variables are used that are always calculated following the MMC connections. This leads to the normalized variables which then subsequently can be used for the next computation step. 

In the result section, first, normalization sub-networks are analyzed individually and the performance for different complexities is given. Secondly, these networks are introduced into the MMC structure and we test the performance and stability of these structures in a series of inverse kinematic tasks.

\subsection{Dynamic Formulation}\label{sect_dynamic}
A further drawback of the classical MMC approach is the characteristic movement profile: the distance towards the target position is decreasing exponentially. In the beginning, this leads to unusual high peaking velocities and, in the latter part, the movement is considerably slowed down. To counter these effects, classical MMC models had been extended and equations containing velocities were introduced \cite{schilling09dynamics}. 

\begin{figure}[tbp]
\centerline{\includegraphics[scale=0.4]{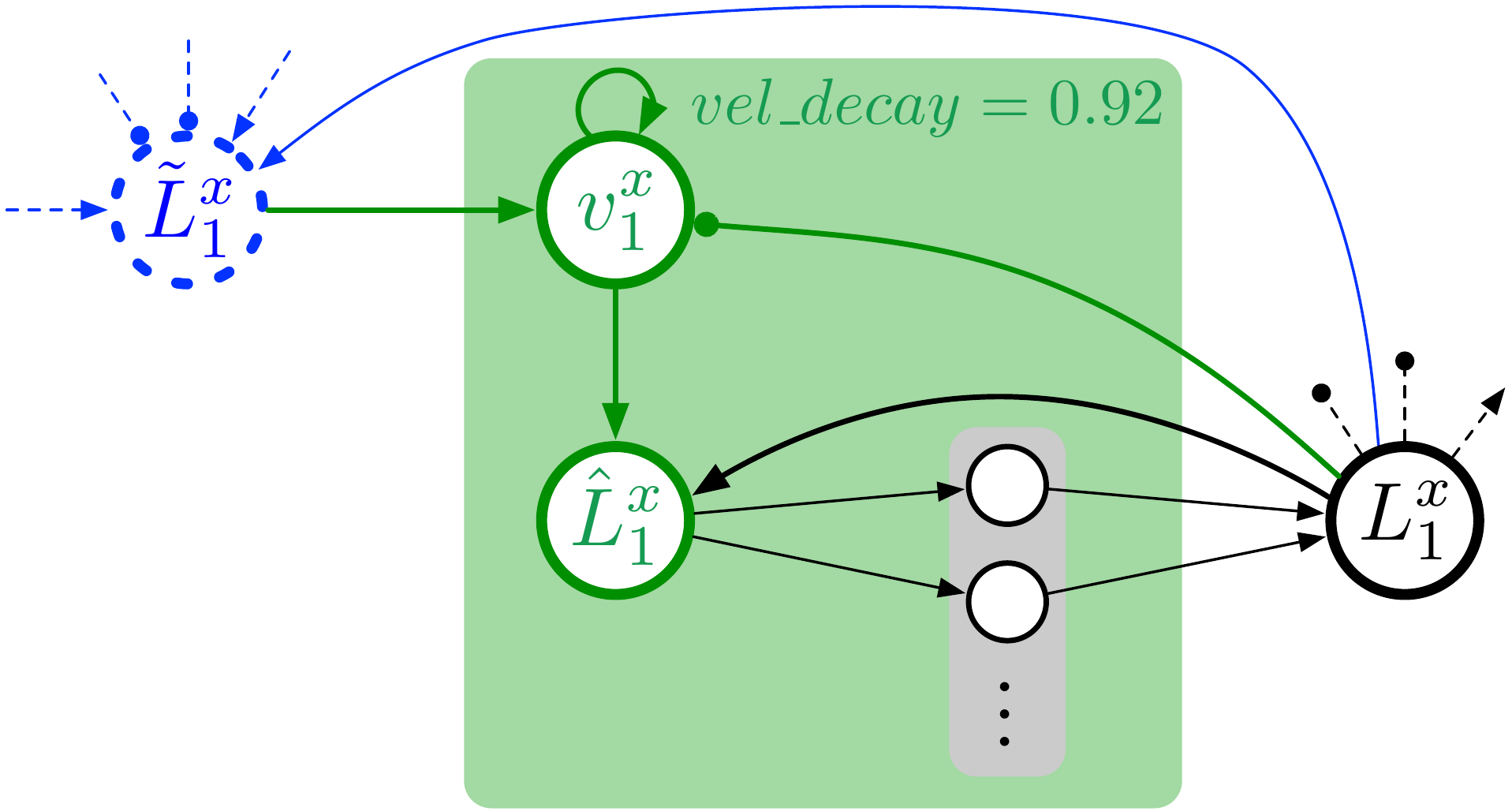}}
\caption{Visualization of the explicit dynamic representation inside the network (in green). Velocities are calculated as the change from the current value to the next assumed value as well as the current velocity. Velocity is used to update the internal representation of the segment vector which afterwards is normalized through the normalization layer (shown in gray).}
\label{fig_dynamic_calculation}
\end{figure}

This dynamic formulation of velocities will be now introduced in the MMC neural network as proposed above. Velocities will be explicitly introduced for the three segment variables. Velocities are calculated simply as the difference of a variable at two different points in time. The introduction of the normalization network already introduced a second variable for each of the segment variables. And, importantly, these two variables relate to different points in time. Internally ($\vec{\tilde {L_i}}$) a variable is already updated towards a new value, depending on the old values (represented in the normalized version $\vec{L_i}$). This makes it straightforward to compute the velocity $\vec{v_i}$ as the difference between those two values (Fig. \ref{fig_dynamic_calculation}). Following the approach in \cite{schilling09dynamics}, the velocity integrates this newly proposed change of the variable as well as the old value of the velocity. These two values are combined as a weighted mean introducing a velocity damping factor (in the following, we chose a velocity damping factor of $d_{vel} = 5$):
\begin{align}
v^x_1 =  \frac{1}{d_{vel}}(\tilde{L}^x_1 - L^x_1) + vel\_decay * \frac{d_{vel} - 1}{d_{vel}}v^x_1
\label{equ_dynamic}
\end{align}

While the general structure of this approach follows \cite{schilling09dynamics}, we apply it here to our fully neural network implementation of a MMC network. First, in order to demonstrate that the dynamic extension works in this specific setup. Secondly, we introduce an additional factor in the equation above---$vel\_decay$. The original dynamic extension acts as a low-pass filter on the internal velocity. When looking at results (see \cite{schilling09dynamics} as well as our results) this lead to overshooting movements which required small return movements after initially passing through the target. The simple introduction of this velocity decay term will proof to remove this problem and lead to much nicer movements (in the sense that velocity is considerably lower and that there is no overshoot). The velocity decay term is employed as a simple multiplicative factor that is smaller than $1$ (in the simulations set to $0.92$). This constantly underestimates the velocity, but the simulations showed that this improved the overall movements (one can consider this as introducing a small friction term that constantly slows down the movement of a segment). The effect of this decay factor is analyzed in detail in the result section.

\begin{figure}[tbp]
\centerline{\includegraphics[scale=0.45]{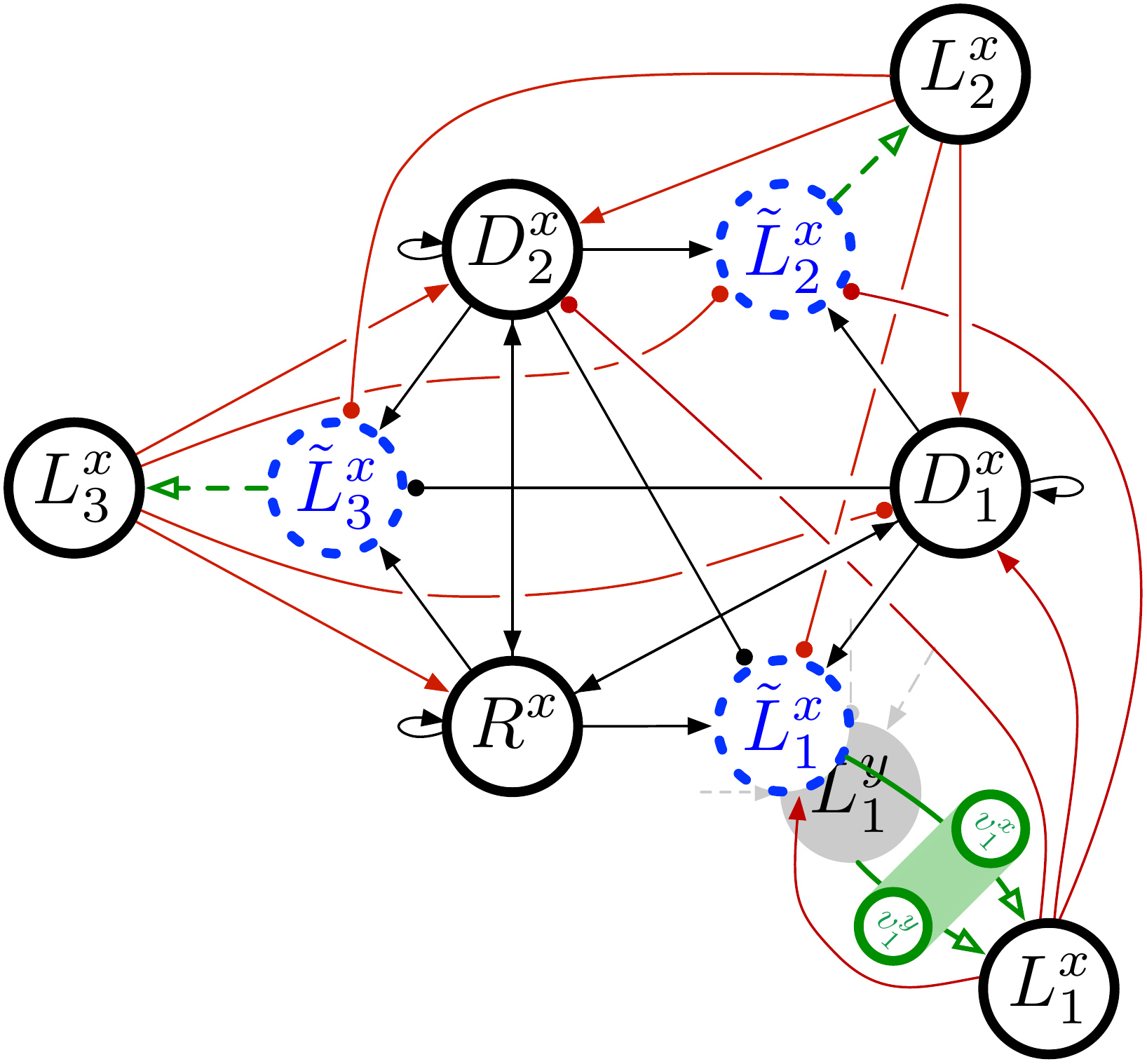}}
\caption{Schematic of the complete recurrent neural network (for the x-component of the vectors). Neurons encode values of the x-component of the vectors. Dashed-blue circles represent the internally updated value after the MMC integration step for the segment vectors. The solid circles ($L_i$) represent the current normalized segment vectors. These are updated depending on the current---internally calculated---velocity followed by the normalization step through the normalization network (shown in green for $L_1$, for the other two segments this is further summarized by the dashed-green arrow).}
\label{fig_dynamic_net}
\end{figure}

The integrated velocity value is now used in a final step---following \cite{schilling09dynamics}---to update the current segment variable $\vec{L_i}$. This update requires the introduction of an additional auxiliary variable $\vec{\hat{L}_i}$ which is equal to the old value of the variable ($\vec{L_i}$) plus the newly internally calculated changed $\vec{v_i}$:
\begin{align}
\vec{\hat{L}_i} =  \vec{L_i} + \vec{v}_1
\label{equ_dynamic_update}
\end{align}
The new internal representation of the segment orientation is used as an input for the normalization neural sub-networks. For the first segment variable the calculation is detailed in Fig. \ref{fig_dynamic_calculation}. The complete extended network is shown in Fig. \ref{fig_dynamic_net}. 

\section{Results}
The presented MMC network was realized for the example of a three-segmented manipulator arm. All three segments were assumed unit length. Implementation was done in Python 2.7.15 using numpy and matplotlib for visualization (all code is available through the GitHub repository https://github.com/malteschilling/normalizationMMC.git ). The normalization layer was realized as a feedforward neural network with hidden layers using keras and tensorflow.  

Results for three different settings will be presented in the following. First, the normalization layer in isolation was tested for different neural network architectures. Secondly, the performance of the MMC network was compared after introducing the normalization sub-networks. And, last, the network including the dynamic extension was tested in a series of inverse kinematics tasks.

\subsection{Learning the Normalization Network}
In a first series of simulation, the capabilities of MLP-based networks for normalization was accessed. A two dimensional input vector served as the input to the network and the network had to provide a two-dimensional output vector. While the input vectors were randomly distorted towards a length in between $0$ and $2$ units, the target for the network was always a vector of the same orientation and unit length. The training data consisted of $3.600$ vectors whose orientation was uniformly distributed around the full circle. For the input vectors, the length was multiplied by a random factor (uniform distribution) from $]0., 2.]$. From this data set $80 \%$ were used as a training set for supervised learning, while the remaining $20 \%$ were used as a test set to access generalization. Mean squared error was employed as a loss. Different feedforward network complexities were tested: the architectures differed with respect to the number of hidden layers and the number of units inside the hidden layers. For the hidden and output units $tanh$ was used as an activation function. All networks were trained using the ADAM optimizer inside the keras framework with a batch size of $16$.

\begin{figure}[tbp]
\centerline{\includegraphics[width=0.9\columnwidth]{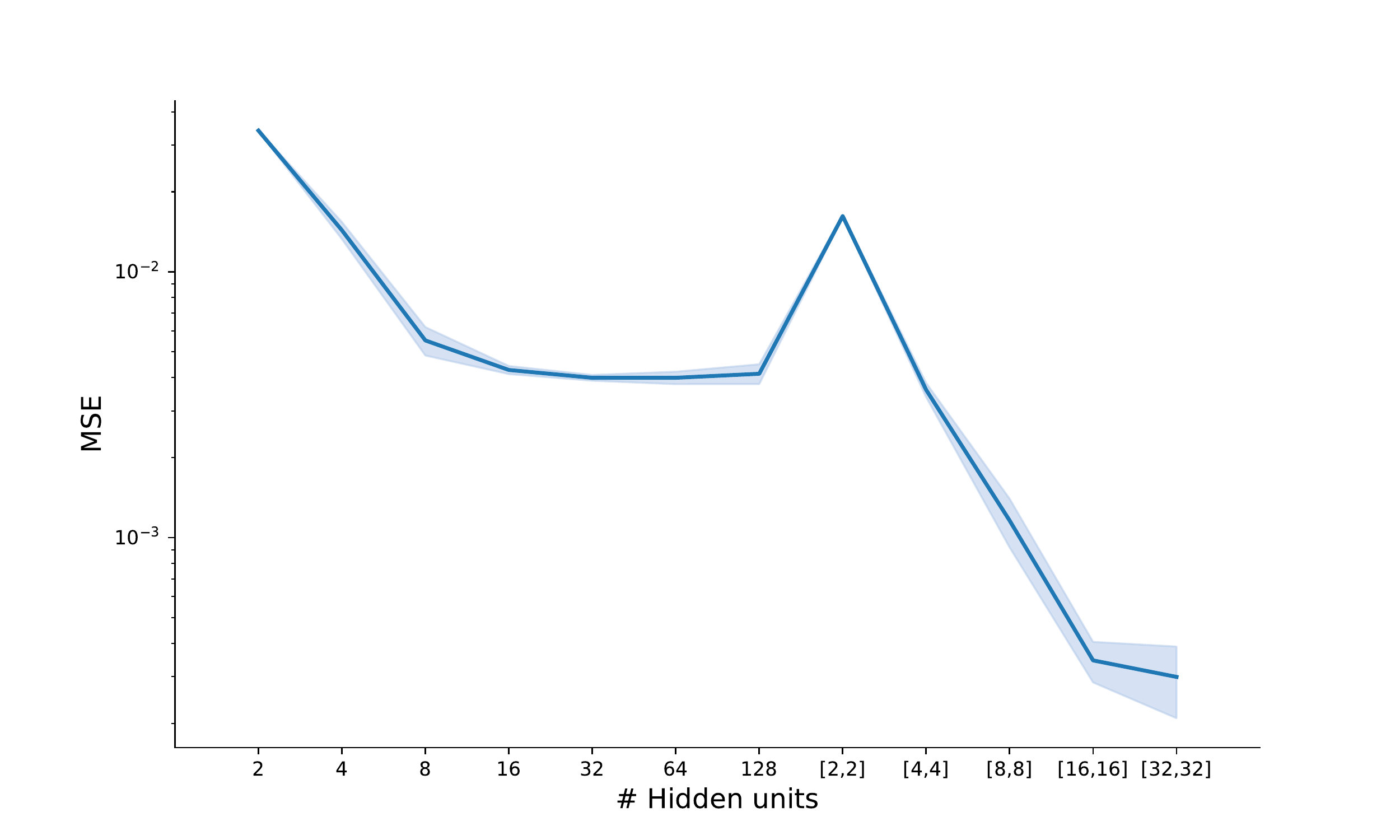}}
\caption{Mean squared error shown for different architectures (different size of a single hidden layer or when using two hidden layers). Solid line shows mean over five simulation runs for each architecture and the shaded area visualizes standard deviation.}
\label{fig_normalization_comp}
\end{figure}

Results are shown in Fig. \ref{fig_normalization_comp}. Given are the mean generalization errors over five simulation runs at the end of training (after 400 epochs the networks settled). Already quite simple architectures with a small number of hidden units lead to a small loss and good performance---for those there is no difference between test and training error (therefore not reported here). The error is quite small and reaches a plateau for an architecture with a single hidden layer (mean squared error for 16 hidden neurons $0.0042$ units). It improves further for a deeper neural network (mean squared error for two hidden layers of each 16 neurons was $0.0003$ units).

\begin{figure}[tb]
\centerline{\includegraphics[scale=0.32]{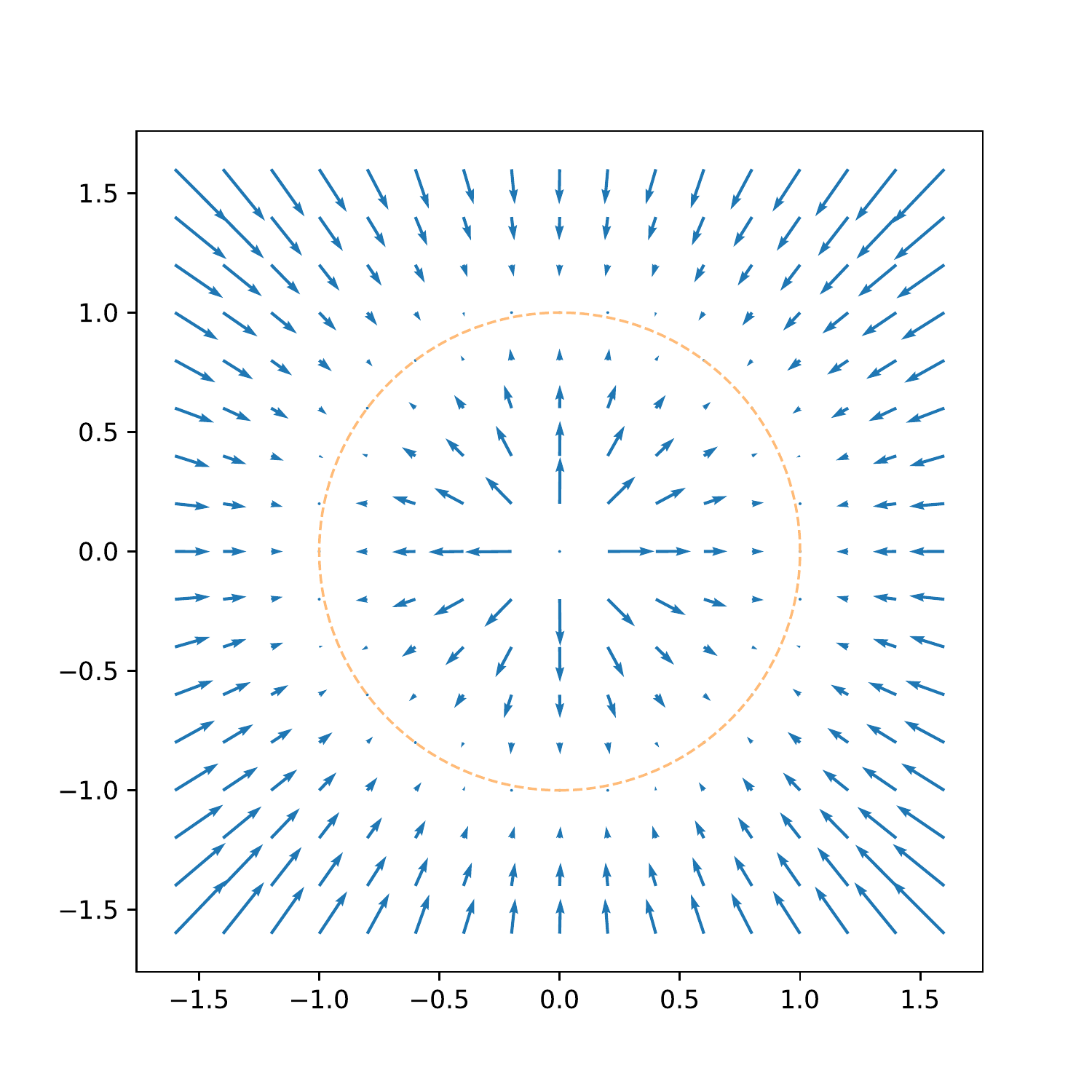}}
\caption{Visualization of the transformation realized by the neuronal network (hidden layer size are eight neurons): over the two dimensional inputs (x and y values) the networks' response is shown as how the respective vectors are changed. The vector arrows show the direction and magnitude of change. As can be seen, the network fulfills its goal moving input positions onto the unit circle.}
\label{fig_norm_transformation}
\end{figure}

Fig. \ref{fig_norm_transformation} visualizes the mapping realized by one of the sub-networks (single hidden layer with $8$ hidden units). Locations in the two-dimensional plane correspond to input values and the blue arrows signify how the inputs are changed by the neural network layer (i.e. these are difference vectors between output and input vectors). As can be seen, all random input vectors are mapped to output vectors on the unit circle.

\begin{figure}[bp]
\centerline{\includegraphics[width=0.9\columnwidth]{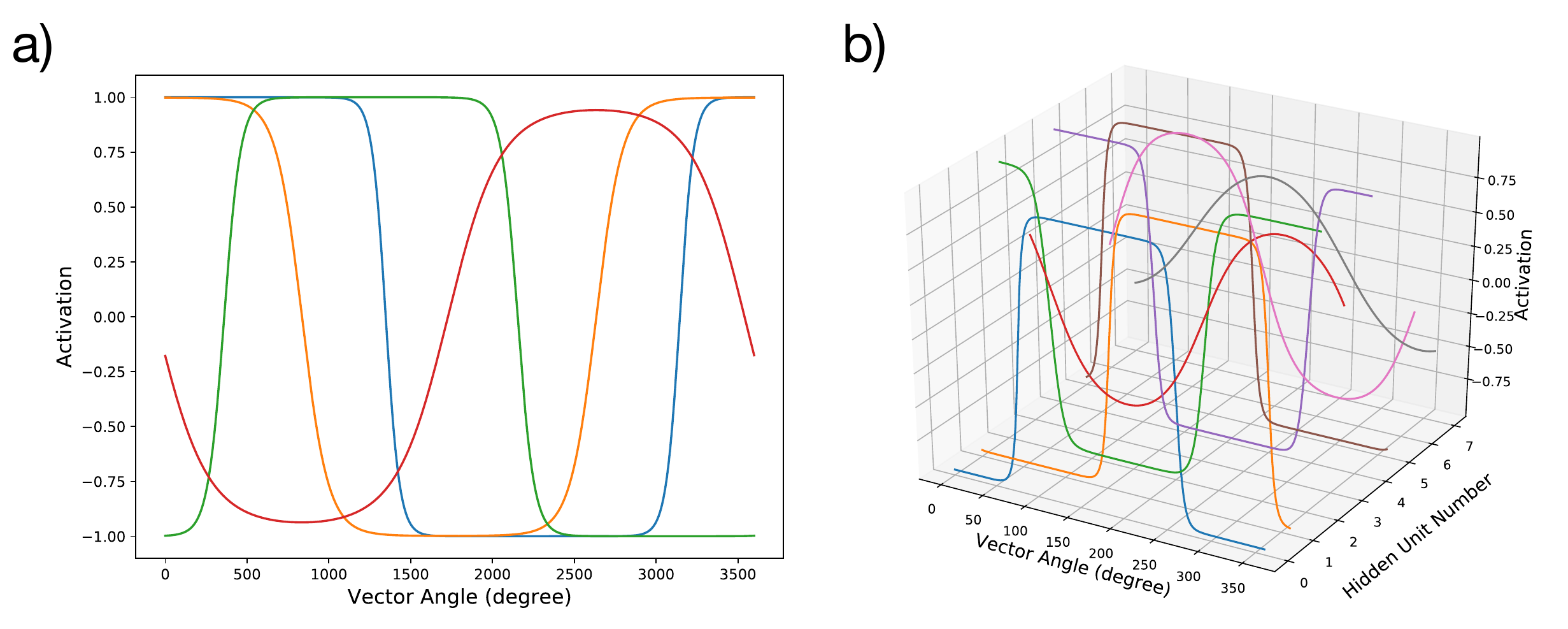}}
\caption{Visualization of the activations of the hidden units (using $tanh$ as an activation function). The x-axis shows the variation of the input (input is a two dimensional vector to the network, the x-axis represents the angle between this vector and the x-axis). Vertical axis is showing how much a unit is active for a given input. In a) this is shown for the case of four hidden units. In b) this is shown in a three dimensional view for 8 hidden units.}
\label{fig_hidden_activation}
\end{figure}

Next, we are looking at the activations of the hidden layers. As two examples, activation of the hidden units for the case of four and eight hidden units are shown in Fig. \ref{fig_hidden_activation}. The hidden units have a prototypical activation that corresponds to a certain range of the orientation of the input vector. This leads to a population-like coding as can be found in place cells or direction cells. Each unit in particular represents a certain part of the input range. With a higher number of hidden units, redundancy had been introduced and there is more overlap. This explains that for a single hidden layer at one point a certain plateau for the error is reached and the network only slightly improves when adding further hidden units. In real neural networks redundancy might help when dealing with noise, but there appears to be little benefit in adding more hidden units in our case.

%


\subsection{MMC network}
In the next step, the normalization network was introduced into the MMC network (using a damping factor of $d=10$) as explained in Sect. \ref{sect_norm_net}. This fully neuronal MMC was analyzed. For the MMC normalization sub-network, first, a network with a single hidden layer and $16$ hidden units was used. In a second step, the performance for the different network complexities from the previous section were compared.

\begin{figure}[tbp]
\centerline{\includegraphics[width=0.7\columnwidth]{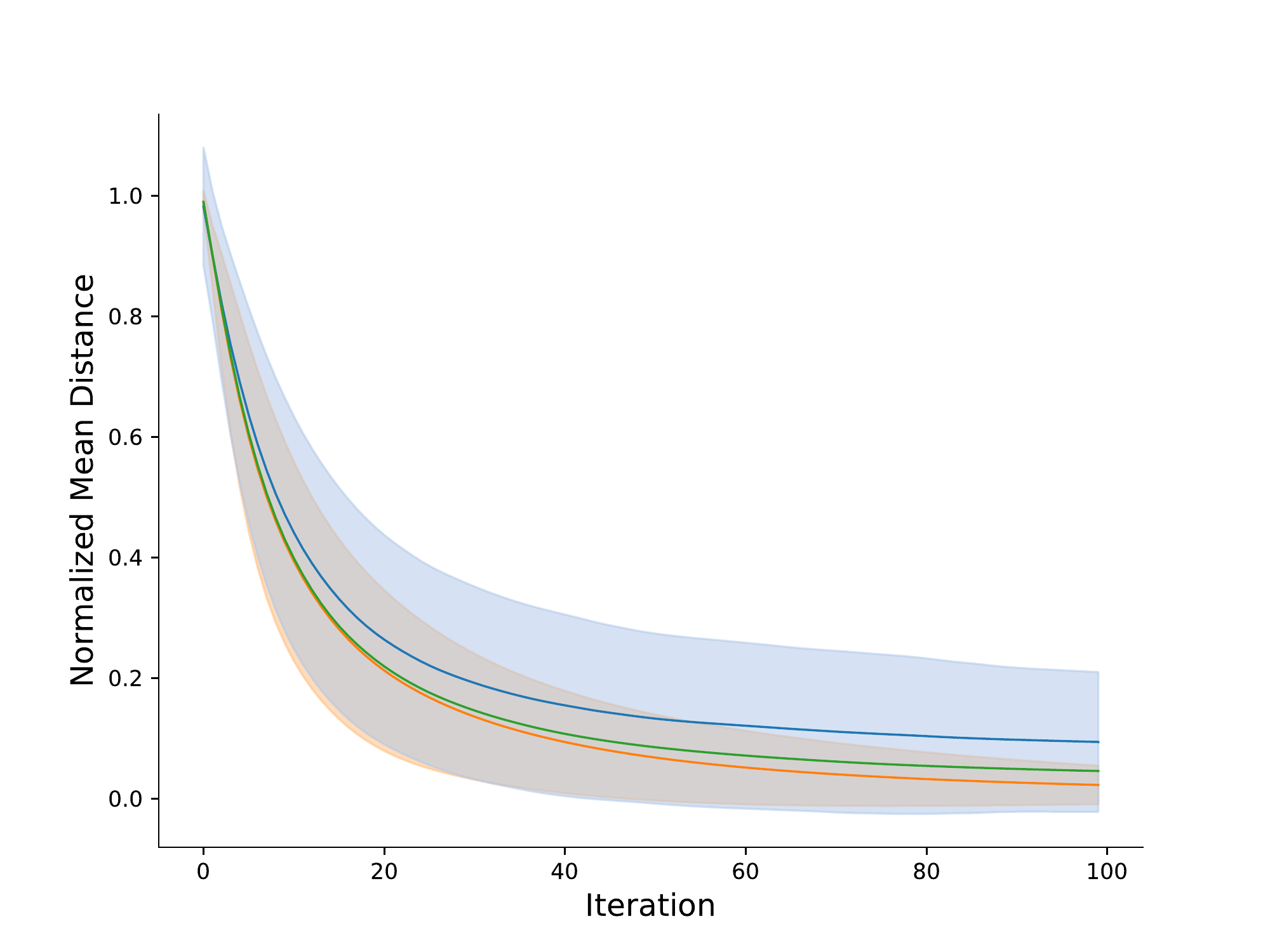}}
\caption{Mean normalized distance to target over time for different normalization methods. As a reference, for the orange curve Euclidean norm was applied on the manipulator segments. Blue curve shows the behavior of the network using a normalization network consisting of a single hidden layer of $16$ hidden units. Shaded areas for these curves represent standard deviation. Green curve shows distance to target for a normalization sub-network with two hidden layers (each 16 hidden units).}
\label{fig_norm_movement}
\end{figure}

The MMC network was tested in a series of inverse kinematic tasks. $21$ points were distributed through half of the whole working space in a systematic manner: three half-circles around the base of the manipulator were used to arrange 7 points on each of these half-circles. The smallest half-circle had a radius of one unit length (or segment length), the second one a radius of two unit length, and the outermost half-circle had a radius of three unit length (which would require the arm to be fully stretched out). The seven points were distributed and equally spaced (every $30$ degrees a point was put onto each half circle). Overall, this lead to $21$ target points and the goal of the MMC network was to make reaching movements starting once from every one of these points towards each of the other points. This resulted in $420$ reaching movements.

Movement of the arm was recorded over all movements. In figure \ref{fig_norm_movement} normalized distance over time towards the target point is shown. The distance is normalized with respect to the distance from start to target point. The comparison shows that the network qualitatively behaves similar to the classical MMC approach and the distance continuously decreased. For the selected complexity of the normalization sub-net the difference between reached and target position is somewhat higher. Standard deviation of the distance is given as the shaded area. Overall, the large variance is due to the fact that many (quite diverse) movements are pooled together from quite different arm configurations. Importantly, both curves show the same trend and the movement profile nicely visualizes the main drawback from classical MMC networks: the distance decreases exponentially which is bad for two reasons. First, with respect to velocity of the manipulator this leads to very high peaking velocities in the first time step. 
Secondly, the movement slows down quickly until it nearly dies down . Therefore the result is not further improving when letting the network run for a longer time. This reasoning motivates the dynamical approach as analyzed later.

\begin{figure}[tbp]
\centerline{\includegraphics[width=0.8\columnwidth]{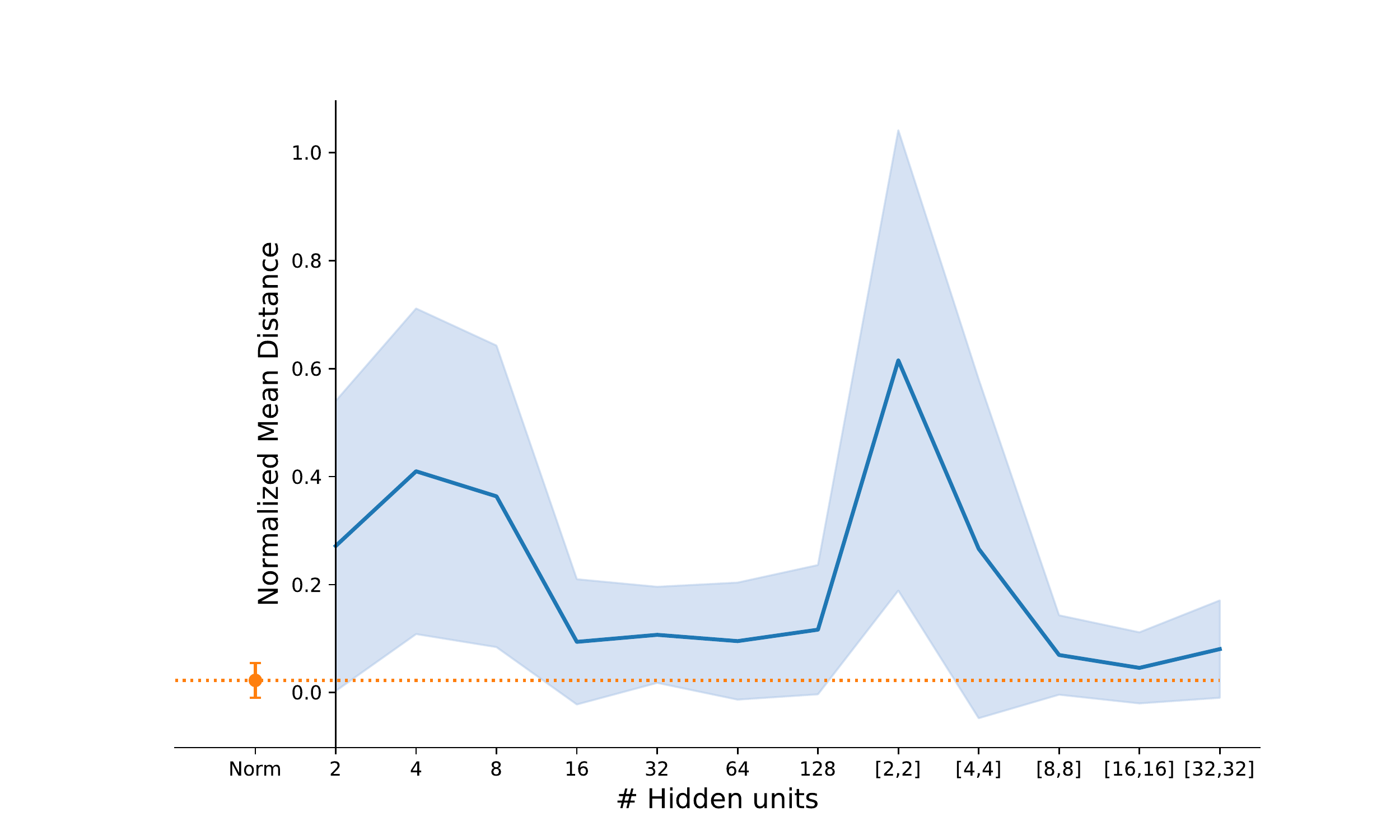}}
\caption{Normalized distance (given on the vertical axis) of end effector to target after a movement of $100$ iteration steps. Comparison shows different network architectures for normalization neural network (shown in blue, shaded area shows standard deviation). As a point of reference, the classical MMC approach using the Euclidean norm for normalization is given in orange.}
\label{fig_norm_network_comp}
\end{figure}

Different normalization network architectures are compared in Fig. \ref{fig_norm_network_comp}. Shown is the normalized distance after $100$ iteration steps for all the MMC networks performing the simulation series described above. Shaded area again represents standard deviation and the dashed orange line provides the classical MMC approach for reference. As found above, the MMC network performs well already for simple architectures consisting of $16$ hidden units in a single layer. When extending the depth of the normalization network to two layers the performance comes close to the performance of the exact classical solution which will therefore be used in the next step.

\subsection{Dynamics}
The dynamic extension of the network was analyzed in further simulations. Fig. \ref{fig_norm_movement} gives an overview of the characteristic temporal behavior of the classical MMC approach as well as when using a normalization network. Distance to the target is decreasing exponentially. Therefore, when considering the velocity profile of such movements this leads to high velocities in the beginning that also decrease exponentially. Such movements are not biologically plausible and the large peak velocities are difficult to realize in real systems.

The MMC network was extended by dynamic equations as explained in section \ref{sect_dynamic} and normalization networks with two hidden layers each consisting of $16$ neurons were employed. A velocity damping factor of $5$ was used (damping factor for the other equations was kept at $d=10$). The same series of repeated reaching movements was recorded as above and compared to results of the classical MMC approach (using Euclidean norm for normalization of segment lengths). 

\begin{figure}[tbp]
\centerline{\includegraphics[width=0.9\columnwidth]{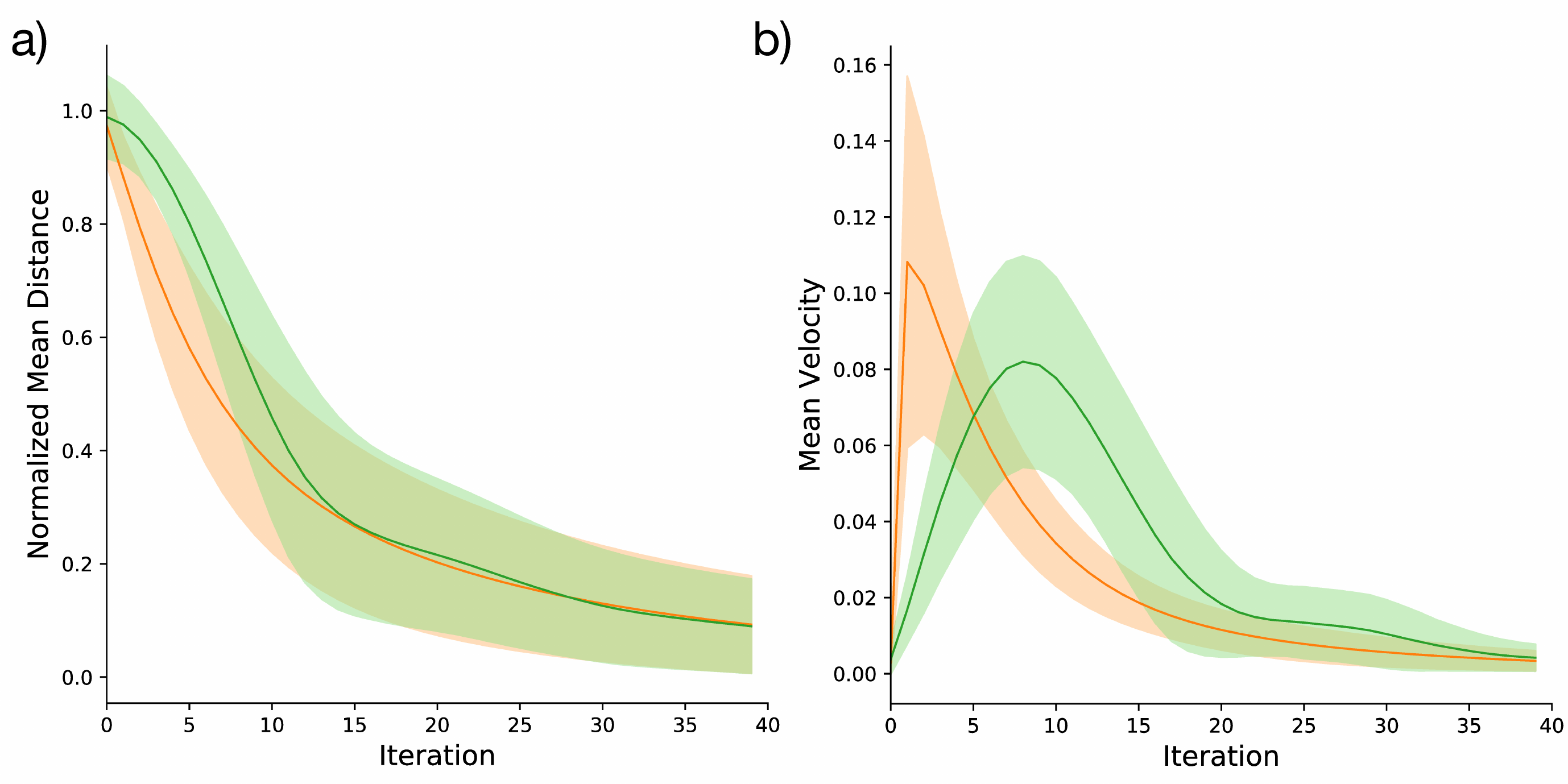}}
\caption{Comparison of classical MMC approach (using Euclidean norm) and dynamic MMC approach (using normalization network with two hidden layers of $16$ neurons each). In a) the normalized mean distance over all $420$ movements is shown over time. In b) velocity profiles are shown (mean velocity is calculated as normalized mean distance per iteration).}
\label{fig_vel_comp}
\end{figure}

Fig. \ref{fig_vel_comp} summarizes the results for all $420$ movements. First, performance was accessed as distance between end effector and target over time. In the early phase, the dynamic MMC lagged the classical approach, but caught up after around $10$ iterations. Both networks reached a similar level after $40$ iterations. Importantly, the velocity profile reflects this behavior as well: while the classical MMC approach shows an initial high peaked velocity, the dynamic MMC network reached a much lower peak velocity and shows a nice bell-shaped velocity profile as characteristic for biological motion.

The mean velocity profile appears to show a second, elongated phase of small to medium velocity between $20$ and $30$ iteration steps. In that period, the velocity was not going further down. When looking at individual movements, we found that the dynamic MMC network tended to overshoot the target and required a return movement. This can be explained as the velocity equations include feeding back the current value of the velocity through the velocity damping factor and in a way introduces low-pass filter characteristics on the velocities of the joints. Therefore, the joints try to maintain a certain velocity even though the MMC network wants to steer them towards lower velocities. Such an effect had been found already for the initial dynamic MMC approach. One solution would be to deal with explicit representation of accelerations \cite{schilling2011dq}, but here we have chosen a much simpler solution and introduced a consistent velocity decay term. It always decreases the velocity by a constant fraction through multiplication with a decay factor (set to $0.92$). This acts like a constant friction as the velocity of all joints is slightly decreased all the time.

\begin{figure}[bp]
\centerline{\includegraphics[width=0.9\columnwidth]{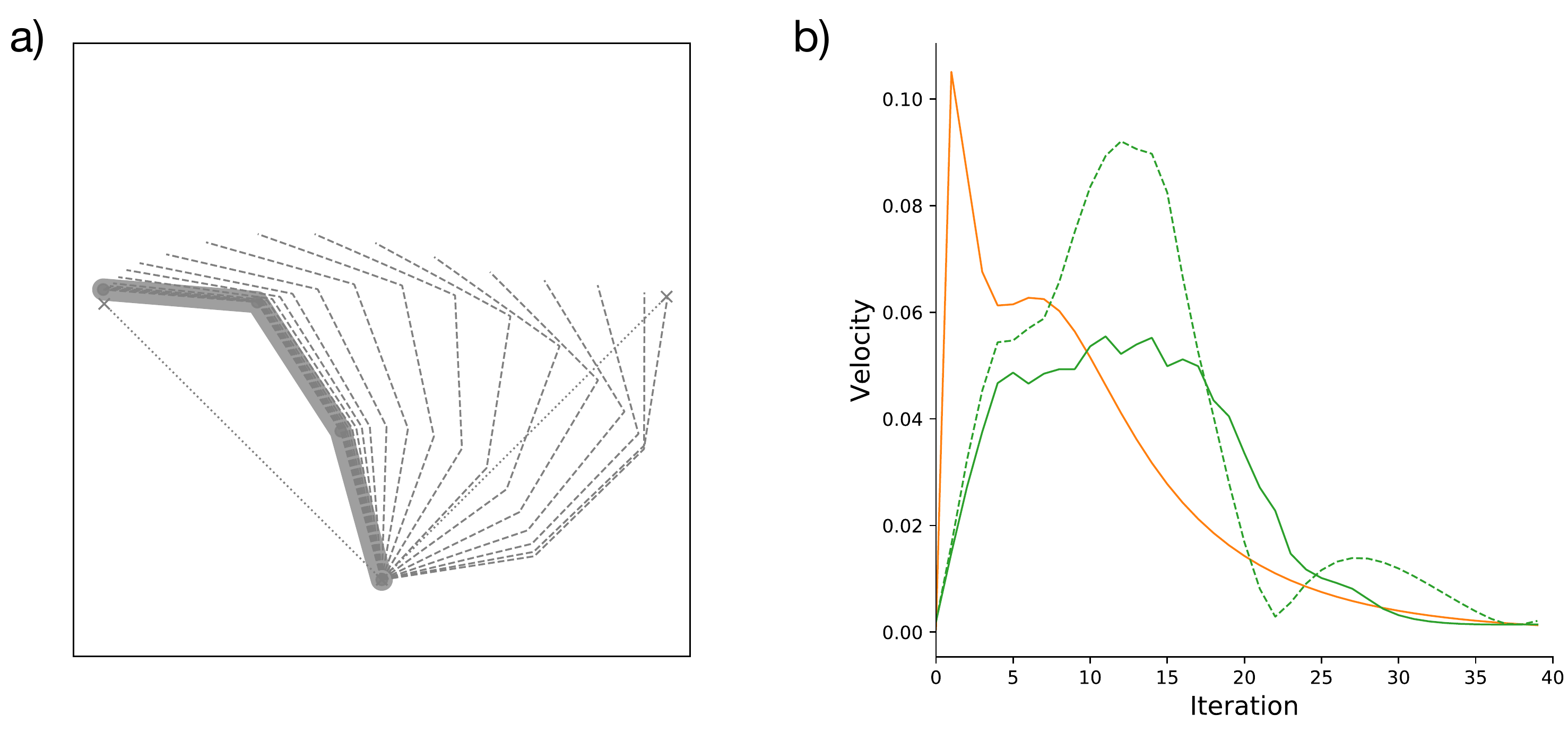}}
\caption{Example for a solution to the inverse kinematic problem using the MMC network with integrated velocities (using a constant decay term for velocities). Shown is the same movement as in Fig. \ref{fig_example_movement} in a). In b) velocity profiles are given for the movements: orange shows velocity for the classical MMC approach and green the extended approach that includes dynamic computations. Solid line represents the final version that includes a permanent decay of velocity, the dashed line shows for comparison velocity profile without this decay term that leads to overshooting of the target (as can be seen by the two peaks in the velocity profiles).}
\label{fig_vel_example}
\end{figure}

The first example movement is shown now for the dynamic MMC network in Fig. \ref{fig_vel_example}. The snapshots of the manipulator illustrate that the movement was much more balanced across the whole space and that there were no huge jumps. Further more, there was no overshoot. The velocity profile for the two dynamic MMC network versions further reveals an important difference. The first version (shown as a dashed green line) shows two velocity peaks, the second one corresponds to an overshoot of the target. In contrast, the second version (solid green line) shows a favorable velocity profile. First, the peak velocity was much lower which is due to the friction-like decay term. Secondly, there was no overshoot which required a return movement. In the end, the version that included a friction term even performed better (normalized distance to target of $0.025$ compared to $0.030$).

\section{Discussion and Conclusion}
A functional internal body model was introduced that is based on the Mean of Multiple Computations principle which breaks down the overall complexity of a controlled system into local relationships. This differs considerably from most approaches to motor control. One system that follows a similar approach is the (neural) Modular Modality Frame (nMMF) model \cite{ehrenfeld2013nMMF} which aims at a probabilistic state estimate based on the integration of redundant information sources. Comparable to our model it is constituted of different frames of reference and transformations between those. One major difference is the focus of the nMMF approach as it used multiple sensory modalities as input and the model is able to integrate these into a consistent state estimate. This allows to deal with imprecise and conflicting sensory information. Sensory information is encoded in a probabilistic way employing population codes. As the MMC approach explicitly deals with integration of redundant information, our approach could be extended towards multiple sensory modalities as well.

But here, we focussed on inverse and forward kinematic as well as dynamic function as required in motor control. The MMC model presented is realized completely as a recurrent neural network. As a novel contribution this required to integrate the normalization that was handled before in an external constraint satisfaction step. The normalization is now realized as a simple feedforward MLP sub-network that keeps the manipulator segments at constant lengths. The results for the internal body model show that this model is able to handle nicely inverse kinematics tasks and, furthermore, due to the dynamic extension is producing actions that show biological movement characteristics. Such a separation of different pathways for kinematics and dynamics is assumed to underlie movement control in animals \cite{MARKOWITZ201844}.

The analysis of the learned normalization layers revealed that these encode limb postures in a prototypical way. While in the nMMF model discussed above a population encoding was explicitly used to represent a posture in a probabilistic way, here a population-based encoding of a posture emerges in the network. This encoding shows place-cell-like activations and fits well to recently described posture cells in rats \cite{mimica_efficient_2018}. Overall, the presented model addresses nicely different aspects on the neuronal organization of functional internal models: First, it shows a population-based encoding which in a next step might be exploited to encode noisy or faulty information in a probabilistic way \cite{baum2015population}. Secondly, the MMC principle is based on redundant frames of reference as are assumed to be used inside the brain. These are, as a third aspect, connected by transformations. While there has been much more research in the area of encodings of spatial information of the environment, findings there point out that there exist redundant forms of representation and transformations which are assumed to happen in the same brain areas \cite{tingley2018transformation}.

\bibliographystyle{IEEEtran}
\bibliography{References}

\end{document}